\documentclass[10pt]{article} 

\usepackage[accepted]{rlj} 

\usepackage{amssymb}            
\usepackage{mathtools}          
\usepackage{mathrsfs}           
\usepackage{graphicx}           
\usepackage{subcaption}         
\usepackage[space]{grffile}     
\usepackage{url}                
\usepackage{lipsum}             

\usepackage{amsmath}
\usepackage{amsthm}
\usepackage{amssymb}
\usepackage{bbm}
\usepackage{mathtools}
\usepackage{mathrsfs}

\usepackage{courier} 

\usepackage{lipsum} 

\usepackage{graphicx}
\usepackage{subcaption}
\usepackage[space]{grffile} 

\usepackage{hyperref}

\usepackage{theoremref}
\usepackage{graphicx}
\usepackage{wrapfig}
\usepackage{xcolor}
\definecolor{dark-red}{rgb}{0.4,0.15,0.15}
\definecolor{dark-blue}{rgb}{0,0,0.7}
\hypersetup{
    colorlinks, linkcolor={dark-blue},
    citecolor={dark-blue}, urlcolor={dark-blue}
}

\usepackage[colorinlistoftodos]{todonotes}

\let\oldnl\nl
\newcommand{\nonl}{\renewcommand{\nl}{\let\nl\oldnl}}

\usepackage{multicol}
\usepackage[utf8]{inputenc} 
\usepackage[T1]{fontenc}    
\usepackage{hyperref}       
\usepackage{url}            
\usepackage{booktabs}       
\usepackage{nicefrac}       
\usepackage[parfill]{parskip}

\usepackage{amsmath,amsfonts,bm}

\def\eqref#1{equation~\ref{#1}}

\def\1{\bm{1}}

\DeclareMathAlphabet{\mathsfit}{\encodingdefault}{\sfdefault}{m}{sl}
\SetMathAlphabet{\mathsfit}{bold}{\encodingdefault}{\sfdefault}{bx}{n}

\def\gS{{\mathcal{S}}}

\newcommand{\E}{\mathbb{E}}

\usepackage{algorithm}

\usepackage{algpseudocode}
\usepackage{xcolor}

\algrenewcommand\algorithmiccomment[1]{\hfill{\color{gray}\footnotesize$\triangleright$ #1}}

\title{Discovering High-Quality Chess Puzzles with Offline Reinforcement Learning}

\setrunningtitle{Discovering High-Quality Chess Puzzles with Offline Reinforcement Learning}

\author{Allen Nie\textsuperscript{1,*}, Anirudhan Badrinath\textsuperscript{1,*}, Nicholas Tomlin\textsuperscript{2,*}, Timothy Dai\textsuperscript{1}, \;\;\;\;\;\;\;\; Carissa Yip\textsuperscript{1}, Rose E Wang\textsuperscript{1}, Emma Brunskill\textsuperscript{1}, Chris Piech\textsuperscript{1}}

\emails{anie@cs.stanford.edu, \ \{ebrun,piech\}@cs.stanford.edu}

\affiliations{
$^{1}$\textbf{Stanford University}\\
$^{2}$\textbf{UC Berkeley}\\
\par 
$^*$ Equal contribution
}

\contribution{
    We formulate chess puzzle recommendation as an offline reinforcement learning problem over a large discrete action space of hundreds of thousands of puzzles and derive an advantage-weighted actor-critic objective with continuous action embeddings that makes policy learning tractable at this scale.
    }
    {
    The actor-critic objective is adapted from~\cite{nair2020awac,kostrikov2021offline}. Continuous action embeddings for large discrete action spaces have been shown to be useful in offline policy evaluation~\citep{saito2022off}. Prior work on offline policy learning in education has operated on substantially smaller datasets and action spaces~\citep{mandel2014offline}.
    }

\contribution{
    We analyze a dataset of over 1.5 billion chess puzzle interactions from 3.1 million users and characterize distinct player learning trajectories, identifying groups whose puzzle Elo improves over time as well as stagnant groups whose Elo remains flat. These differences motivate the need for better puzzle recommendation.
    }
    {
    The dataset was collected from Chess.com over one year. This analysis is descriptive and observational. We do not use player-vs-player game information.
    }

\contribution{
    Through offline policy evaluation on held-out user data, we estimate that the learned policy achieves higher returns than the deployed Chess.com heuristic policy for beginner-level players (puzzle Elo 100--1000), with the largest improvements for the lowest-rated players.
    }
    {
    The evaluation is conducted entirely offline. The policy returns reflect the estimated user's chance of solving the puzzle adjusted by its difficulty after solving previous puzzles, not direct measurement of Elo growth. A complementary LLM-based qualitative analysis using rubrics designed by expert chess players provides preliminary evidence that the learned policy recommends puzzles that are slightly more fun and slightly harder for the target player.
    }


\summary{
Learning and skill mastery require extensive and deliberate practice. In many learning settings, producing high-quality pedagogical materials can require a high level of domain expertise and be very time-consuming. Pedagogical materials often need to train students to engage in different thinking patterns. In some domains, such as chess, puzzles are used to help students practice their skills in calculating the next moves and recognizing known patterns on a board.
Giving students a practice set of puzzles to help them learn different modes of thinking is challenging because the teacher needs to carefully balance between different motifs and how many look-ahead steps a student needs to perform.
Popular online platforms like Chess.com and Lichess offer players millions of puzzles. Unlike chess tactics puzzles procured by human experts, where chess beginners can learn valuable insights, these puzzles are automatically generated and often regarded as having low pedagogical value. These platforms also rely on a heuristic to recommend puzzles to users for practice. Using the user history data over an entire year, we use an offline reinforcement learning algorithm to learn a puzzle recommendation policy and evaluate the policy's effectiveness using offline policy evaluation and expert human rating. 
}

\begin{document}

\maketitle  


\begin{abstract}
Learning and skill mastery require extensive and deliberate practice. In many learning settings, producing high-quality pedagogical materials can require a high level of domain expertise and be very time-consuming. Pedagogical materials often need to train students to engage in different thinking patterns. In some domains, such as chess, puzzles are used to help students practice their skills in calculating the next moves and recognizing known patterns on a board.
Giving students a practice set of puzzles to help them learn different modes of thinking is challenging because the teacher needs to carefully balance between different motifs and how many look-ahead steps a student needs to perform.
Popular online platforms like Chess.com and Lichess offer players millions of puzzles. Unlike chess tactics puzzles procured by human experts, where chess beginners can learn valuable insights, these puzzles are automatically generated and often regarded as having low pedagogical value. These platforms also rely on a heuristic to recommend puzzles to users for practice.
Using the user history data over an entire year, a total of 1.5 billion puzzle-solving histories, we learn the pedagogical value of a puzzle and how to automatically choose a set of puzzles to better support chess learners using insights from offline reinforcement learning.
We show that using offline policy evaluation, our trained policy has significant impact on beginners with puzzle-solving Elo range of 100--1000, particularly for the group of beginners whose learning growth was stagnant.
We also performed a qualitative analysis of the puzzles discovered by our model by collecting annotation ratings from expert chess players.
The success of our pipeline shows promise for a future where we can understand the pedagogical values of practice items given general user interaction data.
\end{abstract}


\section{Introduction}
\label{sec:intro}




Practice makes perfect. The foundation of acquiring knowledge or mastering a new skill relies on countless hours of mindful and deliberate practice~\citep{ericsson1993role,anders2008deliberate}. \cite{koedinger2023astonishing} suggests that across many different learning settings, when students are provided with high-quality, curated learning materials, they can all learn at a similar rate and achieve success with extensive practice. However, the creation of high-quality learning materials is often a key bottleneck. Though lectures can be recorded in video and knowledge can be transcribed in text, students still need to be able to practice what they have learned through forced retrieval and synthesis, which has been shown to greatly enhance learning~\citep{roediger2011ten}. Producing a large amount of practice materials often requires significant human expertise and heavy time investment. 
It is also difficult to evaluate the pedagogical value of each learning material in knowledge acquisition. 

In chess, players often learn by playing against each other directly. However, they also learn important skills through tactics books, which are comprised of puzzles -- subgames limited to a few moves to teach important concepts or to train players to plan multiple moves in a sequence in order to gain more advantage over their opponent. These puzzles are very common for beginners and are considered good learning materials because they isolate and highlight difficult concepts into a small subgame and prime beginners into a habit of thinking strategically~\citep{henkin2021checkmate}. Online chess platforms such as Chess.com and Lichess provide puzzles for players to practice. In order to keep their players engaged, these online platforms aim to serve fresh puzzles on a weekly basis so that players will always have new materials for learning and fun. Chess.com serves roughly 441K puzzles to their players and Lichess hosts over 3.8M puzzles.

In order to produce a large number of new puzzles quickly, unlike classic tactics books where puzzles are curated by human experts through careful deliberation, review, and editing, online chess platforms use an algorithm to automatically generate puzzles from the actual games played by players. 
The puzzles are assigned an Elo rating and presented to players at random, provided their difficulty is within a fixed range of the player's rating~\citep{chesscom2023puzzledatabase}. However, it remains unclear whether these puzzles effectively contribute to a player’s long-term chess learning. In a system where learning materials are automatically generated, it is essential to filter out content that fails to produce meaningful, lasting gains in knowledge or skill.


Offline reinforcement learning (RL) can learn a policy from a dataset of historical interactions. It has been used to discover patient treatment policies in the ICU~\citep{komorowski2018artificial,luo2024position}, to create personalized learning paths in math education software~\citep{mandel2014offline,bassen2020reinforcement,ruan2024reinforcement}, and to learn new controllers for robotics~\citep{kumar2020conservative,kumar2021workflow,lu2022challenges}. It has been shown to scale with data and can learn policies that generalize beyond the training distribution~\citep{kumar2022offline}. 

Using a large dataset of 1.5 billion puzzle solving attempts over the course of a year from Chess.com with 3.1 million users, we can use an offline RL algorithm to learn a policy and evaluate its effectiveness using a holdout portion of the dataset. We first show that, similar to the finding in \cite{wang2022surrogate}, we can separate users into two groups: a growth group, where the user's Elo rating gradually increases with more puzzles, and a stagnant group, where the user's Elo remains flat. We then propose a simple advantage-weighted actor-critic objective for offline policy learning based on~\cite{nair2020awac,kostrikov2021offline} and use continuous action embeddings to account for the large action space. Finally, we show that, estimated with one-step importance sampling, the policy learned to serve puzzles significantly more effectively for beginners (Elo\footnote{Throughout this paper, we use ``rating'', ``Elo'' to refer to the puzzle Elo score established by Chess.com: \url{https://www.chess.com/leaderboard/tactics}. Both user and puzzles are assigned an Elo score. Both change initially, but after a while, puzzle's Elo score becomes fixed and ceases to change. We do not use any information outside of tactics.} 100--1000) than the original Chess.com system. Our qualitative evaluation provides preliminary evidence that the learned policy recommends puzzles that may be slightly more fun and somewhat harder relative to the user's rating.




%

\section{Related Work}
\label{sec:related}

There is a long history of using games such as chess to evaluate the progress of AI~\citep{campbell2002deep,silver2017mastering,silver2018general}. Although these models have developed superhuman abilities to master the game, few investigations have focused on leveraging their knowledge to teach humans. \cite{schut2023bridging} and \cite{mcgrath2022acquisition} did pioneering work on uncovering chess knowledge learned in AlphaGo and AlphaZero. \cite{mcilroy2021detecting} built models to identify the chess playing styles of each user and then later released models that imitate chess players of different levels~\citep{mcilroy2022learning}. \cite{hamade2024designing} built a chess agent that can match a weaker player's skill to foster skill-compatible learning. However, no end-to-end system has been developed to recommend chess puzzles specifically for teaching humans.

In automated teaching systems, reinforcement learning has been used to adaptively select instructional materials for students. \cite{chi2009elicit} built an adaptive tutoring system for simple decision making in college math education. \cite{ruan2024reinforcement} built a math tutoring chatbot that decides when and how to provide hints. \cite{mandel2014offline} uses offline RL to augment the fraction learning experience of students in math games. \cite{nie2023understanding} examines how RL-based personalization in math tutoring systems differentially impacts subgroups of students. \cite{liu2022giving} leverages meta-exploration to give feedback on interactive student programs. However, most of these projects remain small in scale, and offline RL has not demonstrated usefulness for large-scale educational settings. In our work, we follow \cite{kumar2022offline} to scale offline RL models to large datasets with billions of interactions and use offline RL to discover high-quality puzzles that can improve a chess player's learning experience.


\section{Data}
\label{sec:data}

Our dataset consists of players' puzzle history data from 3,132,428 unique, active users of a popular chess website, Chess.com, playing a total of 1,536,254,297 puzzles (441,113 unique puzzles) over the course of one year from March 2021 to March 2022. On average, a user played 490.4 puzzles over this one year. Of those 490.4 puzzles, 96.9\% are played within three minutes of another puzzle, which indicates that players tend to play puzzles in short bursts of time. An average burst comprises 5.1 games. Playing in bursts naturally suggests that users do not continuously play puzzles throughout the day. In fact, the average time between these bursts is approximately two and a half days. In general, our dataset reflects the engaged and extensive user base of online chess players.


\paragraph{Puzzle Serving} 

According to Chess.com administrators, the site serves puzzles to users using a bucketed uniform policy. Both puzzles and users are assigned an Elo score, specifically Chess.com's tactics/puzzle-solving rating rather than a rating from games against other humans (see footnote in Section \ref{sec:intro}). This policy serves a player's next puzzle by first bucketing all chess puzzles that have puzzle Elo ratings within $\pm 200$ points of a player's Elo and then sampling uniformly from that bucket. As a player begins to fail puzzles, that $\pm 200$ bucket eases to $-300$/$+100$ after one incorrect puzzle, $-400$/$+0$ after two, $-500$/$-100$ after three, and $-600$/$-200$ after four incorrect puzzles. In this way, the Chess.com policy adapts to player performance by serving progressively easier puzzles in response to a player's difficulty in solving previously served puzzles. Overall, Chess.com's \textit{bucketed uniform policy} integrates a player's performance on their past four puzzles, their Elo rating, and puzzle rating to determine the next puzzle served.

\paragraph{Chess Puzzles}

Chess puzzles are core units of learning. In a chess puzzle, the player is presented with an initial position and tasked with finding the correct solution in one or more moves, adhering to the puzzle's main goal or task, described in part through a puzzle's motifs. The initial position is shown to the player as a board, and it can also be conveniently encoded as a FEN (Forsyth-Edwards Notation) string. Move counts for puzzles in our dataset range from 1 to 16, averaging at 2.6. Harder puzzles tend to demand more moves, with the puzzle's difficulty reflected in its puzzle rating. Our puzzles' ratings range from 100 to 4000. In contrast to move count and puzzle rating, which are both numerical, motifs provide more qualitative insights into the diversity of puzzles.

A puzzle's motifs describe the primary strategy or skill required to solve the puzzle. Examples of such motifs include ``Exchange Sacrifice'' and ``Promotion,'' describing a notable event that takes place in a puzzle. A puzzle can have more than one motif; in fact, the puzzles in our dataset have, on average, 5.6 motifs each, with motif count ranging from no motifs at all to 56 motifs. Of our 441,113 puzzles, 26\% do not have any motifs at all. Thus, motifs provide informative yet nonexhaustive insights into the quality and variety of served puzzles.



\begin{figure}[h]
    \centering
    \begin{subfigure}[t]{0.47\textwidth}
        \centering
        \includegraphics[width=\textwidth]{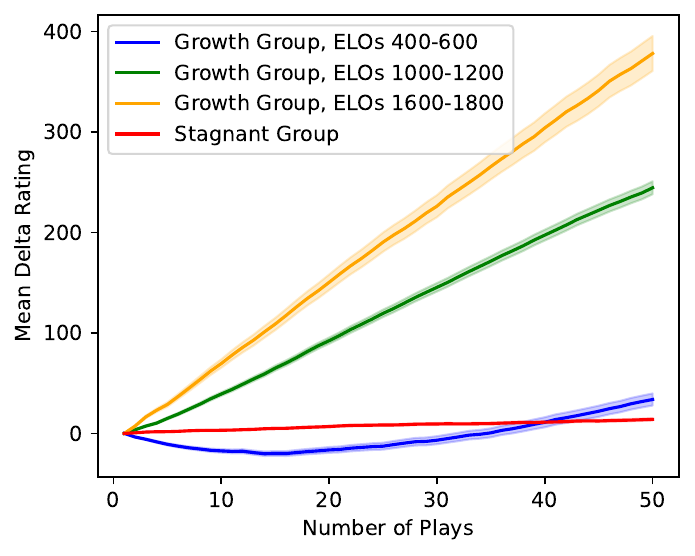}
        \caption{Plot illustrating the mean change in rating for four groups of players based on their Elo ranges and their relative increases in Elos.}
        \label{subfig:line_num_plays_v_mean_delta_rating}
    \end{subfigure}
    \hfill
    \begin{subfigure}[t]{0.49\textwidth}
        \centering
        \includegraphics[width=\textwidth]{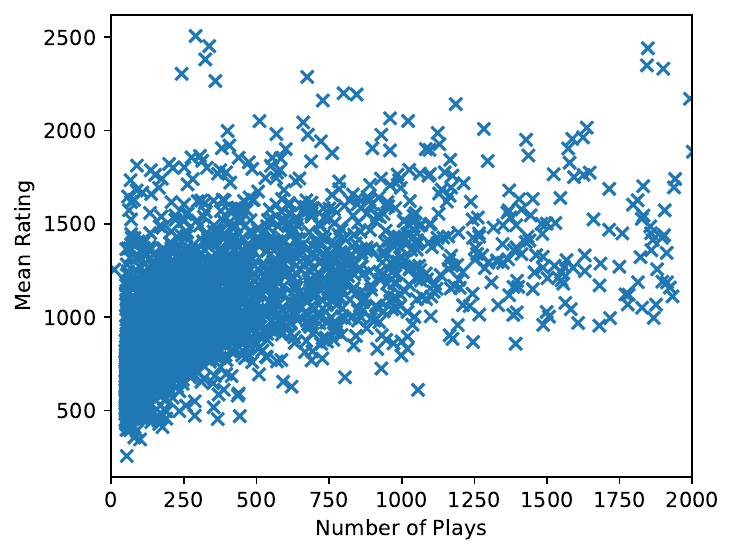}
        \caption{Scatter plot showing the relationship between a player's puzzle play count and their rating averaged over a year.}
        \label{subfig:scatter_num_plays_v_mean_rating}
    \end{subfigure}
    \caption{Plots that illustrate user Elo growth, showcasing the Elo growth trajectories of several groups of players (Figure \ref{subfig:line_num_plays_v_mean_delta_rating}) alongside overall Elo trends across our entire player base (Figure \ref{subfig:scatter_num_plays_v_mean_rating}).}
    \label{fig:analyzing_user_Elo_growth}
\end{figure}

\paragraph{Analyzing User Elo Growth} In Figure \ref{subfig:line_num_plays_v_mean_delta_rating}, we plot the mean change in rating across the first 50 recorded puzzle plays for four groups of players based on their Elo ranges and relative increases in Elos. The ``Growth Group'' represents players in the 99th percentile for Elo increase (the change in Elo from each player's first recorded Elo) among players in their Elo range, while the ``Stagnant Group'' represents players in the 1st percentile for Elo increase, combined across all three Elo ranges. The trajectory of the Stagnant Group, as expected, is relatively flat, signifying little growth in Elo across the first 50 games played. On the other hand, players in the Growth Group of high Elos improve rapidly in their first 50 games, with less rapid growth for lower Elo players, and even less rapid growth for the lowest Elo players. This suggests that a player's average, long-term Elo generally points to their initial improvement speed. Interestingly, the Growth Group of the lowest Elo players experiences a dip in Elo in their first $\sim$40 games, falling below Elo improvements of even the Stagnant Group, before seeing gains. This dipping trend suggests that, for overall inexperienced players, playing chess puzzles involves an especially difficult acclimatization period, during which unfamiliarity with the game may initially hinder performance, but players eventually learn the rules and complexities of the game and reap the benefits of playing more puzzles.

We broaden our analysis in Figure \ref{subfig:scatter_num_plays_v_mean_rating} and examine players' total puzzle counts. Each point in the figure represents an individual player's total number of puzzles played. We observe a general trend from the bottom left toward the upper right, indicating that a higher number of puzzles played generally correlates with a higher Elo. This strengthens the suggestion that playing chess puzzles is a skill that improves with more experience. The total number of puzzles played is generally an indicator of a player's skill level.

\section{Policy Learning}
\label{sec:method}


\subsection{Preliminaries}

We define a stochastic decision process $M =$ $\langle \gS, A, \mathcal{T}, r, \gamma \rangle$, where $\gS$ is a set of states; $A$ is a set of discrete actions; $\mathcal{T}$ is the transition dynamics; $r$ is the reward function; and $\gamma \in (0, 1)$ is the discount factor. Let $\mathcal{D} = \{\tau_i\}_{i=1}^n$, where $\tau = \{(s_j, a_j, s_j', r_j)\}_{j=0}^{H-1}$, with $s_j' \equiv s_{j+1}$, denote a dataset of trajectories collected under some policy on $M$ with time horizon $H$. We denote the performance of a policy $\pi$ based on its expected discounted return $R_t = \E_{\tau \sim \rho_\pi}[\sum_{t'=t}^{H-1} \gamma^{t'-t} r_{t'}]$ where $\rho_\pi$ is the distribution of $\tau$ under policy $\pi$. By the definition of the action-value and value function respectively, we let $Q(s, a) = \E_{\tau \sim \rho_\pi}[R_t \mid s, a]$ and $V(s) = \E_{\tau \sim \rho_\pi}[R_t \mid s]$.

In an off-policy policy learning problem, we do not have access to the true transition dynamics $\mathcal{T}$ or any online interaction with $M$. Instead, we take an offline dataset $\mathcal{D}$, which can be collected by one or a group of distinct policies, which we collectively refer to as the behavior policy $\pi_b$ on the decision process $M$.

In actor-critic frameworks, we perform policy evaluation and policy improvement in conjunction to derive an optimal policy $\pi^*$ \citep{konda1999actor}. Typically, policy evaluation is performed using iterative application of the Bellman update. In the context of deep RL for off-policy learning (i.e., where we have $\mathcal{D}$, policy $\pi_\theta$ parameterized by $\theta$, action-value function parameterized by $\phi$), we perform policy improvement through gradient updates to the policy $\pi_\theta$ and optimize Equation~\ref{eqn:basic}.
\begin{align}
    \arg \max_\theta \E_{\mathbf{s} \sim \mathcal{D}, \mathbf{a} \sim \pi_\theta (\cdot \mid \mathbf{s})}[Q^\pi_\phi(\mathbf{s}, \mathbf{a})]
    \label{eqn:basic}
\end{align}
For our formulation, we consider a non-Markovian policy $\pi_\theta$ based on a transformer architecture, where the state and action space incorporate a fixed length context. Specifically, our state space consists of a continuous embedding-based $k$-dimensional representation of the user's puzzle history and learning progress (e.g., Elo), i.e., $s \in \mathcal{S} = \mathbb{R}^k$. Our desired action space is the set of all $N = 441{,}113$ chess puzzles, i.e., $a \in \mathcal{A} = \{1, 2, \dots, N\}$.

In an off-policy setting, common pitfalls with traditional actor-critic techniques include extrapolation error by venturing outside of the supported data distribution in $\mathcal{D}$, which results in an accumulation of errors from bootstrapped action-value functions \citep{sutton2018reinforcement}. To ensure that the trained policy $\pi_\theta$ stays close to the behavior policy, we penalize the statewise Kullback-Leibler divergence $D_{\mathrm{KL}}\!\left(\pi_\theta(\cdot \mid \mathbf{s}) \,\|\, \pi_b(\cdot \mid \mathbf{s})\right)$, averaged over states sampled from $\mathcal{D}$, with coefficient $\beta$ \citep{rafailov2024direct,tutor2016conservative}.
\begin{align}
        &\arg \max_\theta \E_{\mathbf{s} \sim \mathcal{D}}\!\left[
        \E_{\mathbf{a} \sim \pi_\theta(\cdot \mid \mathbf{s})}\!\left[Q^\pi_\phi(\mathbf{s}, \mathbf{a})\right]
        - \beta D_{\mathrm{KL}}\!\left(\pi_\theta(\cdot \mid \mathbf{s}) \,\|\, \pi_b(\cdot \mid \mathbf{s})\right)\right]
    \label{eqn:base}
\end{align}
\subsection{Deriving an Offline Policy Learning Objective}

We derive a simple advantage-weighted actor-critic-style objective for offline policy learning, based primarily on \cite{nair2020awac} and \cite{kostrikov2021offline}. We leverage offline advantage estimation via a parameterized value function $V^\pi_\psi(\mathbf{s})$, augmenting the objective shown in Equation \ref{eqn:base} with a baseline. Importantly, note that since the value function and its inputs are constants with respect to the optimization variable, $\theta$, it does not bias or modify the objective.
\begin{align}
\arg \max_\theta \E_{\mathbf{s} \sim \mathcal{D}}\!\left[
\E_{\mathbf{a} \sim \pi_\theta(\cdot \mid \mathbf{s})}\!\left[Q^\pi_\phi(\mathbf{s}, \mathbf{a}) - V^\pi_\psi(\mathbf{s})\right]
- \beta D_{\mathrm{KL}}\!\left(\pi_\theta(\cdot \mid \mathbf{s}) \,\|\, \pi_b(\cdot \mid \mathbf{s})\right)\right]
\end{align}

Based on the derivation in \cite{nair2020awac}, we can project the closed-form optimal solution $\pi^*(\mathbf{a} | \mathbf{s}) \propto \pi_b(\mathbf{a} | \mathbf{s}) \ \exp(\frac{1}{\beta} (Q^\pi_\phi(\mathbf{s}, \mathbf{a}) - V^\pi_\psi(\mathbf{s})))$ into the policy space by minimizing $D_{\mathrm{KL}}\!\left(\pi^*(\cdot \mid \mathbf{s}) \,\|\, \pi_\theta(\cdot \mid \mathbf{s})\right)$ over the state distribution, which yields the weighted maximum likelihood objective shown in Equation \ref{eqn:test2}.
\begin{align}
    L_\pi(\theta) = \mathbb{E}_{\mathcal{D}}[-\log \pi_\theta (\mathbf{a} | \mathbf{s}) \exp (\frac{1}{\beta}(Q^\pi_\phi(\mathbf{s}, \mathbf{a}) - V^\pi_\psi(\mathbf{s})))]
    \label{eqn:test2}
\end{align}

Mirroring \cite{kostrikov2021offline}, we train the value network and action-value network for advantage estimation purely using transitions in the dataset $\mathcal{D}$. To train the value network, we leverage expectile regression, as shown in Equation \ref{eqn:value}, and to train the action-value network, we use the objective shown in Equation \ref{eqn:actionvalue}.
\begin{align}
    L_V(\psi) = \mathbb{E}_{\mathcal{D}} [L_2^\tau(Q^\pi_\phi(\mathbf{s}, \mathbf{a}) - V^\pi_\psi(\mathbf{s}))] \label{eqn:value}\\
    L_Q(\phi) = \mathbb{E}_{\mathcal{D}} [(r(\mathbf{s}, \mathbf{a}) + \gamma V^\pi_\psi(\mathbf{s}') - Q^\pi_\phi(\mathbf{s}, \mathbf{a}))^2] \label{eqn:actionvalue}
\end{align}

\subsection{Architecture for Chess Puzzle Recommendation}

\begin{figure}
    \centering
    \includegraphics[width=\linewidth]{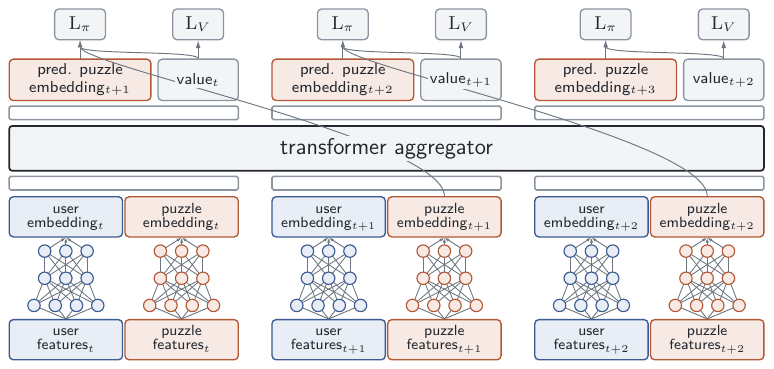}
    \caption{Architecture and training procedure of the chess puzzle recommendation policy, leveraging user features and puzzle features to optimize a policy-based loss $L_\pi$ and a value function via $L_V$.}
    \label{fig:enter-label}
\end{figure}
 
 Since we formulate the recommendation problem as an offline policy learning problem, we use a causal transformer architecture as our decision making policy, denoted by $T_{\theta_T}$. 
We choose this architecture because it is widely used for student knowledge tracing and modeling in education~\citep{pandey2019self,choi2020towards,shin2020saint+}.
At each sequence element (i.e., time step $t$), the transformer takes as input the concatenation of a user representation $u_t$ and a puzzle representation $p_t$, each constructed by applying a neural network to the corresponding user and puzzle features. The outputs at each sequence element are (a) a prediction of the next puzzle, $\pi(\mathbf{a} \mid \mathbf{s})$, and (b) a value function estimate, $V(s)$. 

During training, we optimize $L_\pi$, $L_Q$, $L_V$ on predictions at each sequence element, with training sequences sampled uniformly from the chess puzzle interaction data. Importantly, we apply teacher forcing during the forward pass of training, which is standard in training autoregressive decoder-only transformers.

\paragraph{User Embedder} To construct a latent representation of the user, we embed user-specific features $f_u$, i.e., the user Elo and user correctness at time $t$, using a small, two layer multi-layer perceptron (MLP), denoted as $U_{\theta_U}$. We normalize the output user embedding $u_t = U_{\theta_U}(f_u)$ to unit norm (e.g., L2 normalization).

\paragraph{Puzzle Embedder} To construct a latent representation of the puzzle, we embed puzzle-specific features $f_p$ using an MLP and convolutional neural network (CNN), denoted as $P_{\theta_P}$. Specifically, we use a CNN to encode the board position associated with the puzzle, alongside a learned embedding for each individual puzzle and its first move. Similarly to the user embeddings, we normalize the puzzle embedding $p_t = P_{\theta_P}(f_p)$ to unit norm (e.g., L2 normalization).

\paragraph{Practical Considerations for Policy Learning} Given a large discrete action space $\mathcal{A}$ (i.e., approximately half a million puzzles), computing and normalizing the policy probabilities $\pi_\theta(a \mid s)$ across the entire action space is intractable during training. We therefore parameterize the policy using an exponentiated inner product between the transformer's output and each candidate puzzle embedding. Let $h_t = T_{\theta_T}(\mathbf{p}_{t-T:t}, \mathbf{u}_{t-T:t})$ denote the transformer's output for the history through time $t$, and let $p_a = P_{\theta_P}(f_a)$ denote the embedding of candidate puzzle $a$. Equation~\ref{eqn:final} defines the resulting temperature-weighted softmax policy.
\begin{align}
\pi_\theta(a \mid s_t)
= \frac{\exp\!\left(\lambda h_t^\top p_a\right)}
{\sum_{a' \in \mathcal{A}} \exp\!\left(\lambda h_t^\top p_{a'}\right)}
\label{eqn:final}
\end{align} 
Although puzzle embeddings can be precomputed, evaluating the denominator in Equation~\ref{eqn:final} over the entire action space remains expensive during training. We therefore approximate it using the puzzle embeddings $P(B)$ in minibatch $B$ as candidate negatives. For the observed next puzzle $a_{t+1}$ with embedding $p_{t+1}$, the approximation is
\begin{align}
\hat{\pi}_\theta(a_{t+1} \mid s_t)
= \frac{\exp\!\left(\lambda h_t^\top p_{t+1}\right)}
{\sum_{p_n \in P(B)} \exp\!\left(\lambda h_t^\top p_n\right)}
\label{eqn:final2}
\end{align} 
Afterwards, we apply the losses $L_\pi$, $L_Q$ and $L_V$ to optimize the policy with respect to the transformer parameters and user and puzzle embedders. We summarize the entirety of the training algorithm in Algorithm \ref{algorithm:supervised_learning} and as depicted in Figure \ref{fig:enter-label}.

\begin{algorithm}
  \caption{Offline training algorithm for chess puzzle recommendation}
  \label{algorithm:supervised_learning}
  \begin{algorithmic}
    \State \textbf{Input:} $\mathcal{D},\; T_{\theta_T},\; U_{\theta_U},\; P_{\theta_P}$
    \For{each minibatch $B \subset \mathcal{D}$}
        \State $f_p, f_u \gets B$, $p_t \gets P_{\theta_P}(f_p), u_t \gets U_{\theta_U}(f_u)$ \Comment{puzzle and user features}
        \State $\hat{\pi}_\theta(\mathbf{a}_{t+1} \mid \mathbf{s}_t)
        \gets
        \frac{\exp\!\left(\lambda\, T_{\theta_T}(\mathbf{p}_{t-T:t}, \mathbf{u}_{t-T:t})^\top p_{t+1}\right)}
        {\sum_{p_n \in P(B)} \exp\!\left(\lambda\, T_{\theta_T}(\mathbf{p}_{t-T:t}, \mathbf{u}_{t-T:t})^\top p_n\right)}$
        \Comment{in-batch approximation}
        \State $L_V(\psi) \gets \mathbb{E}_{\mathcal{D}}\!\left[L_2^\tau\!\left(Q^\pi_\phi(\mathbf{s},\mathbf{a}) - V^\pi_\psi(\mathbf{s})\right)\right]$
        \State $L_Q(\phi) \gets \mathbb{E}_{\mathcal{D}}\!\left[\left(r(\mathbf{s},\mathbf{a}) + \gamma V^\pi_\psi(\mathbf{s}') - Q^\pi_\phi(\mathbf{s},\mathbf{a})\right)^2\right]$
        \State $L_\pi(\theta) \gets \mathbb{E}_{\mathcal{D}}\!\left[-\log \hat{\pi}_\theta(\mathbf{a}\mid\mathbf{s})
        \exp\!\left(\frac{1}{\beta}\left(Q^\pi_\phi(\mathbf{s},\mathbf{a}) - V^\pi_\psi(\mathbf{s})\right)\right)\right]$
        \State $(\theta_T,\theta_U,\theta_P,\phi,\psi) \gets
(\theta_T,\theta_U,\theta_P,\phi,\psi)
- \eta \nabla (L_V + L_Q + L_\pi)$
    \EndFor
  \end{algorithmic}
\end{algorithm}

\subsection{Deriving a Reward Function}


During evaluation, we compute puzzle embeddings once for the entire action space in order to obtain the policy's probability distribution over puzzles. This is an upfront cost rather than a per-query one: the same puzzle embeddings are reused across all user sequences and do not need to be recomputed for each new prediction.

We construct a scalar reward function that attempts to quantify the learning benefit in terms of the user's correctness on the puzzle and  relative puzzle difficulty. Trivially, a correct response on a challenging puzzle is ideal, and it highlights progression in terms of learning, warranting a large reward. In any cases where the user answers incorrectly, it demonstrates no verifiable evidence of learning, so we assign no reward in these cases. Otherwise, in cases of partial correctness, we assign rewards based on the proportion of correct moves and relative difficulty of the puzzle.

\begin{wrapfigure}{r}{0.43\textwidth}
\vspace{-24pt}
  \begin{center}
    \includegraphics[scale=0.32]{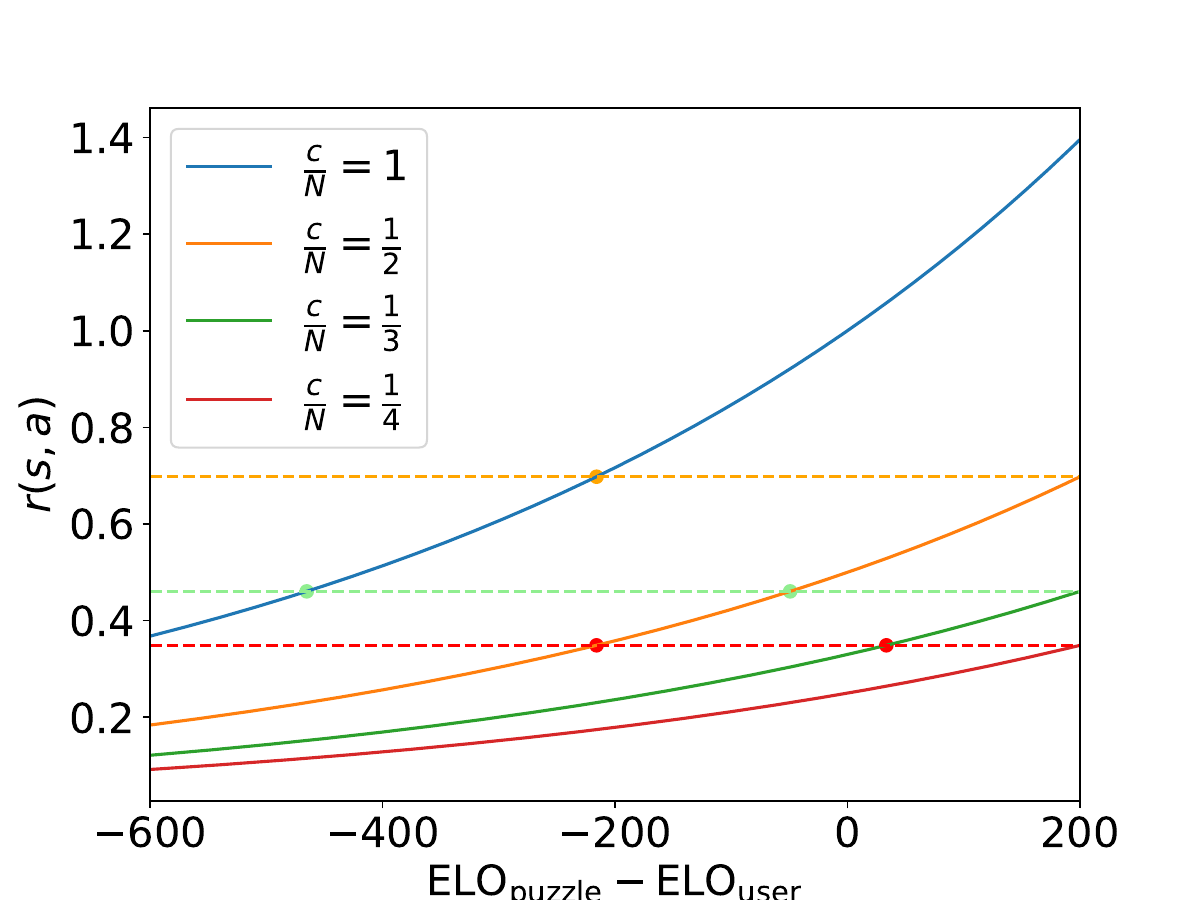}
  \end{center}
    \caption{\small Reward function $r(s, a)$ for varying levels of correctness and Elo differentials between puzzle and user, with reward equivalences shown between balances of correctness and relative difficulty. }
    \label{fig:gradew2}
    \vspace{-10pt}
\end{wrapfigure}

Given the number of total moves $N$, number of correct moves (from the user) $c$, and the Elo of the puzzle and user, we assign the reward based on the difficulty-weighted correctness, with the difficulty weight $\alpha$ being a hyperparameter.
\begin{align}
 r(s, a) = \frac{c}{N} \exp(\alpha \cdot (\mathrm{Elo}_\mathrm{puzzle} - \mathrm{Elo}_\mathrm{user}))
\end{align}

We provide a visualization of the reward function across different proportions of correct moves $c / N$ in Figure \ref{fig:gradew2} for $\alpha = 0.002$, which is the chosen hyperparameter due to its reasonable balance between partial correctness and difficulty.  Specifically, based on internal testing, $\alpha = 0.002$ provides a reasonable level of reward equivalence between partial correctness levels (e.g., halfway correct) on challenging puzzles and complete correctness on easier puzzles (e.g., puzzles rated 200--400 Elo points lower). These equivalencies are displayed via dotted lines, where the color corresponds to the maximal reward achieved at a particular correctness level, and the intersection with other lines indicates the Elo delta at the reward equivalence.

\section{Evaluation}
\label{sec:eval}

While it would be ideal to run a randomized controlled trial experiment with the learned policy, there are many practical challenges. It is hard to recruit enough players to participate in the experiment, and it may take a significant time to be able to measure any difference reliably. We instead use a mixture of offline policy evaluation estimators and human evaluation to understand whether the learned policy could have meaningful impact on learning.

\subsection{Evaluation Using Importance Sampling}

To evaluate our trained policy with respect to the behavior policy, we can leverage importance sampling, using the known probability distributions of the behavior policy (e.g., as deployed by Chess.com) and our trained policy. Importantly, traditional importance sampling with sequences of size 256 (or even a fraction of that) is impractical due to the significant induced variance in the estimator over large sequences. Instead, we employ a one-step approximation (e.g., removing past weights) across 32 steps, which significantly reduces variance at the cost of being biased \citep{chen2019top}. We refer to this as one-step importance sampling. To further reduce variance, we clip the importance weight for each step $\pi_\theta(\mathbf{a} | \mathbf{s}) / \pi_b (\mathbf{a} | \mathbf{s})$ between $\epsilon_s = 0.1$ and $\frac{1}{\epsilon_s} = 10.0$. 


To quantify our improvements over the behavior policy for different users, we break down our results into different Elo groups. Additionally, we examine groups of stagnant growth and different growth groups, similarly to Figure \ref{fig:analyzing_user_Elo_growth}; for this analysis, we use a simple definition of stagnant growth, which is growth of no more than 50 Elo points across the sequence. Importantly, this allows for examination of our improvements on different types of users and learning patterns, e.g., improvements on stagnant or new users may be more critical to retention on the Chess.com platform.    

\subsection{Expert Evaluation}

\paragraph{Annotation Rubrics} To annotate the quality of chess puzzles, we designed a detailed rubric with two chess experts involved in the project (USCF rating around 1900 and 2400). The rubric covers different characteristics of a puzzle, from whether it tests the player's ability to perform in-depth calculations of sequential moves or it focuses on testing the ability to recognize iconic patterns. In addition, the experts are also asked to judge if the puzzle is appropriate for a player of a given rating, whether it is a high-quality puzzle, and if it is fun to play. The rubric took multiple iterations with internal testing to verify its robustness and is described in full in Appendix~\ref{app:annotation_guidelines}.

\begin{wrapfigure}{h}{0.3\textwidth}
\vspace{-5mm}
\begin{tabular*}{0.3\textwidth}{@{}lc@{}} 
\toprule
Category                                                       & Range \\ \midrule
Calculation                                                    & 1--4   \\
Pattern          & 1--3   \\
Informativeness                                                & 1--5   \\
Appropriateness                                                & 1--5   \\
Quality                                                        & 1--5   \\
Fun                                                            & 1--5  \\
\bottomrule
\end{tabular*}
\caption{Qualitative rating categories designed in collaboration with chess expert consultants.}
\vspace{-5mm}
\label{fig:rating_category}
\end{wrapfigure}

We recruited 8 chess experts to annotate 30 puzzles. The annotators were eight strong chess players, 5 of whom hold official chess titles. In total, there are 2 USCF Experts, 1 USCF National Master, 1 FIDE Master, 1 FIDE International Master, and 2 FIDE Grandmasters. The chess annotators' USCF Elo ratings are between 1990 and 2576. For a rough (not directly comparable) point of reference, the chess legend Magnus Carlsen has a FIDE rating of 2832, and the 100th highest-rated player has a FIDE rating of 2634 as of Spring 2025.

The puzzles are randomly sampled from a \textit{bucketed uniform policy} based on the player's rating range, and we have 6 players with different Elo scores. We chose 12 puzzles as \textit{preference learning} puzzles that are annotated by all experts. This means that out of the 30 puzzles each expert annotated, only 18 puzzles are unique between them. The annotation was performed through an anonymized spreadsheet. Each annotator was asked to spend approximately 1.5 hours on the annotation and was later compensated with a gift card.

\paragraph{Scaling Annotations with LLMs}
Judging whether a chess puzzle is high quality or not requires annotators with significant expertise, who are distinctly different from standard crowd workers. This also means we are not able to perform massive annotations that can take up to hundreds of hours. However, sometimes, two policies that learn to recommend puzzles can behave very similarly and, therefore, require a lot of annotations to understand if they are statistically significantly different. To help with scaling up the annotation effort, inspired by \cite{zheng2023judging,he2023annollm}, we use the expert annotations to calibrate a large language model (LLM) that can imitate an expert chess annotator's judgments. LLMs seem to be able to understand FEN strings and can play chess with some deliberation~\citep{toshniwal2022chess,feng2023chessgpt}. Without making sweeping generalizations, we can leverage LLMs to provide preliminary understanding of whether the chess puzzles chosen by two systems are different.

With example expert annotations, we use DSPy~\citep{khattab2023dspy} to create our  LLM chess puzzle judge. DSPy is an LLM library that allows us to specify a set of initial prompts, training data, and a metric. It then automatically selects the best expert examples to include in the LLM prompt that can maximize the metric. Internally, DSPy performs cross-validation with a random search to find the best subset of expert annotations. 
We learn 8 LLM annotators using the data from 8 chess annotators. We separately learn 8 annotators because each annotator's preference might be different, and imitating the behavior of 8 individuals seems more difficult than imitating the behavior of one individual. For each annotator, we use the 12 shared \textit{preference learning} puzzle annotations described above, splitting them evenly into 6 training and 6 validation examples for DSPy prompt optimization.


\paragraph{Prompt Design} We use a system message and the DSPy Signature module to design the prompt. The system message specifies the persona of the annotator using their USCF Elo rating and title. The DSPy signature contains three input fields---the annotation guidelines, the target user's Chess.com puzzle Elo, and the puzzle board---and one output field containing the six category scores. Appendix~\ref{app:llm_prompt_fields} provides a complete description of these fields. For puzzle board representation, we experimented with a FEN string and a grid-like representation of the entire board. We found the grid-like representation to work better. Along with the board position, we also describe the first move of the puzzle in text.  






\section{Experimental Results}

\paragraph{Training}
To train our chess puzzle recommendation policy, the dataset is partitioned randomly by users, with 90\% used for training and 10\% for evaluation. During training, each batch consists of 16 examples of users' puzzle histories, with a sampled sequence length of 256 to accommodate computational constraints. We use a discount rate of 0.99. During training, to implement Equation~\ref{eqn:final2}, we use 65K puzzles as the in-batch negative examples to estimate our policy's probabilities. 
Due to the large dataset size, we train the policy for 25K steps, alongside the learned action-value and value functions.



In this section, we examine the performance of our learned policy in two different ways. We use an offline policy evaluation method to provide an objective comparison between our policy's performance and the behavior policy's performance. Then, we use our automated annotation pipeline to provide another set of ratings.

\subsection{Policy Evaluation}

\begin{figure}
    \centering
    \includegraphics[width=0.53\linewidth]{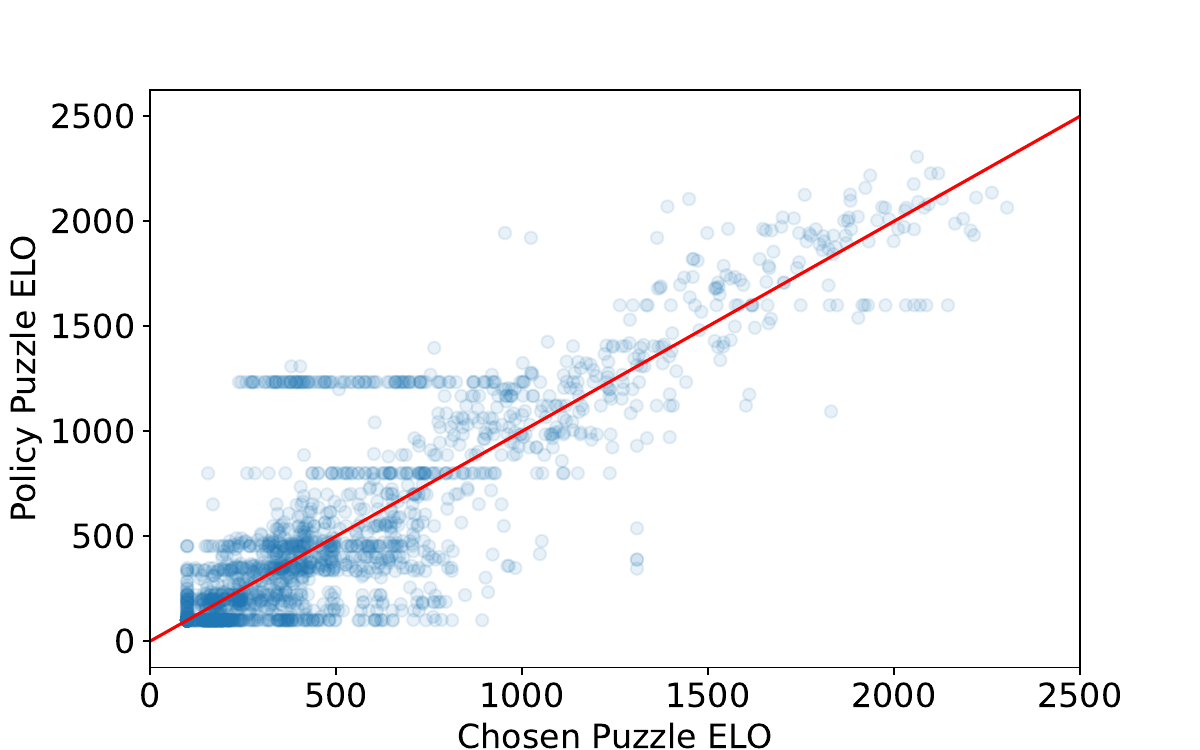}\includegraphics[width=0.53\linewidth]{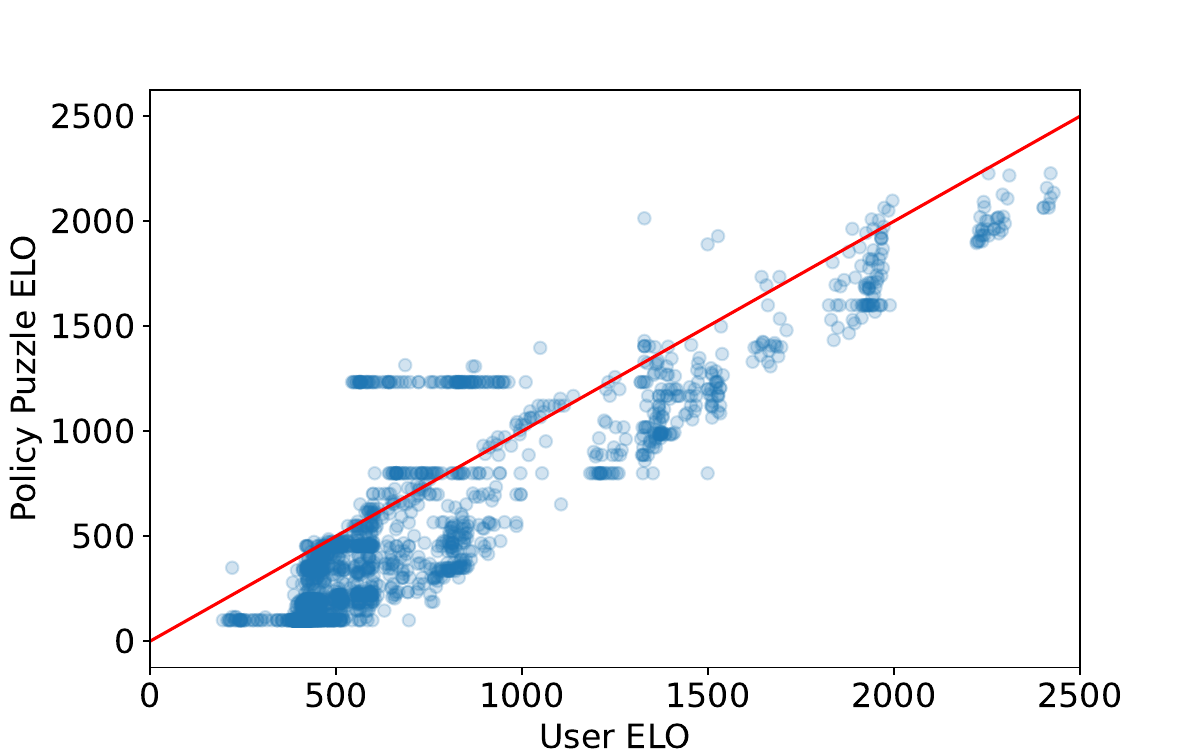}
    \caption{Distribution of trained policy's top recommended puzzle versus the puzzle Elo of the chosen puzzle (left) and the user Elo (right).}
    \label{fig:enter-label3}
\end{figure}

To demonstrate that our policy performs well qualitatively and quantitatively relative to the behavior policy, we perform evaluations based on importance sampling and visualize and summarize the policy recommendations. As a preliminary qualitative evaluation, we examine the distribution of puzzle Elos recommended by our policy with respect to the puzzle chosen by Chess.com's current policy in Figure \ref{fig:enter-label3}. As a preliminary sanity check, our policy consistently recommends puzzles in the rough ballpark of the chosen puzzle and the user Elos, though our policy tends to recommend harder puzzles on average, especially for users with higher Elo. An interesting trend is that the recommendations typically underestimate the user Elo, which reflects the behavior of the behavior policy as well.

\begin{table}[ht!]
    \centering  
        \vspace{1em}
\begin{tabular}{c || c c | c c || c c}    
\toprule      
\textbf{Elo} & \multicolumn{2}{c|}{\textbf{Stagnant Group}} & \multicolumn{2}{c||}{\textbf{Growth Group}}  & \multicolumn{2}{c}{\textbf{Average}} \\      
       & $\pi_b$  & $\pi_\theta$ & $\pi_b$  & $\pi_\theta$ &  $\pi_b$  & $\pi_\theta$   \\    
\hline 
100--600     & $14.3 \, \text{\scriptsize $\pm$} \, \text{\scriptsize 1.5}$ & $\mathbf{52.9} \, \text{\scriptsize $\pm$} \, \text{\scriptsize \textbf{3.0}}$ & ${17.7} \, \text{\scriptsize $\pm$} \, \text{\scriptsize {0.4}}$ & $\mathbf{58.8} \, \text{\scriptsize $\pm$} \, \text{\scriptsize \textbf{1.7}}$ & $16.9 \, \text{\scriptsize $\pm$} \, \text{\scriptsize 0.5}$ & $\mathbf{57.4} \, \text{\scriptsize $\pm$} \, \text{\scriptsize \textbf{1.5}}$ \\ 
600--1000     & $14.1 \, \text{\scriptsize $\pm$} \, \text{\scriptsize 1.2}$ & $\mathbf{27.0} \, \text{\scriptsize $\pm$} \, \text{\scriptsize \textbf{6.7}}$ & ${20.9} \, \text{\scriptsize $\pm$} \, \text{\scriptsize {0.5}}$ & $\mathbf{42.3} \, \text{\scriptsize $\pm$} \, \text{\scriptsize \textbf{3.2}}$ & $19.5 \, \text{\scriptsize $\pm$} \, \text{\scriptsize 0.4}$ & $\mathbf{39.2} \, \text{\scriptsize $\pm$} \, \text{\scriptsize \textbf{2.9}}$ \\ 
1000--1500     & $12.3 \, \text{\scriptsize $\pm$} \, \text{\scriptsize 0.4}$ & $12.1 \, \text{\scriptsize $\pm$} \, \text{\scriptsize 1.2}$ & $18.3 \, \text{\scriptsize $\pm$} \, \text{\scriptsize 0.6}$ & $19.7 \, \text{\scriptsize $\pm$} \, \text{\scriptsize 2.4}$ & $15.3 \, \text{\scriptsize $\pm$} \, \text{\scriptsize 0.4}$ & $16.0 \, \text{\scriptsize $\pm$} \, \text{\scriptsize 1.4}$  \\ 
1500+     & $12.2 \, \text{\scriptsize $\pm$} \, \text{\scriptsize 0.2}$ & $12.2 \, \text{\scriptsize $\pm$} \, \text{\scriptsize 1.2}$ & ${15.9} \, \text{\scriptsize $\pm$} \, \text{\scriptsize {0.3}}$ & $16.1 \, \text{\scriptsize $\pm$} \, \text{\scriptsize 1.1}$ & $13.5 \, \text{\scriptsize $\pm$} \, \text{\scriptsize 0.2}$ & $13.6 \, \text{\scriptsize $\pm$} \, \text{\scriptsize 0.9}$ \\ 
\bottomrule    
\end{tabular}   
\caption{Returns obtained by trained policy $\pi_\theta$ and the Chess.com policy $\pi_b$ through one-step importance sampling. We show 95\% confidence intervals with $\pm$ and boldfaced numbers are statistically significant with Student's t-test.} 
\label{tab:performance_comparison}  
\end{table}

In Table \ref{tab:performance_comparison}, we summarize the results of the one-step importance sampling approaches compared to the behavior policy for different Elo buckets and growth groups. Across lower Elo buckets (between 100 and 1000), we show statistically significant and consistent improvement compared to the behavior policy. However, notably, as the Elo increases, our margins of improvement decrease, where our policy is neutral with respect to the behavior policy.

Qualitatively, this is sensible because optimal puzzle selection is more critical to learning at earlier stages, whereas missteps in puzzle recommendation may not lead to large outcome differences for an experienced chess player. Additionally, an important note is that the vast majority of the user base (and reflected in the training and evaluation data) is concentrated at lower Elo levels, with over 80\% of users under 1500 Elo, where our improvement is roughly 110.1\% over the existing system. 

While the return is consistently lower for the stagnant group (as is expected), the relative performance improvement from the trained policy is approximately similar across growth groups and stagnant groups. That said, for the smallest Elo bucket, $\pi_\theta$ performs better relatively on stagnant group users, whereas for all others, $\pi_\theta$ performs relatively better on growth group users.


\subsection{Expert Evaluation}



We provide a qualitative analysis using the expert-designed rubrics and with the help of an LLM to understand whether there is any noticeable difference between the type of puzzles recommended by our learned policy and the original Chess.com puzzle serving system. We randomly sample 200 users from the holdout evaluation dataset. To evaluate the original Chess.com policy $\pi_b$, we directly use what was actually recommended to the user at that time.  For the trained model $\pi_\theta$,  we sample the top-2 puzzles that have the highest probabilities of being recommended.


\begin{table}[h]
\centering
\small
\begin{tabular}{r||cc|cccc}
\toprule
Method & Calculation & Pattern Recognition & Fun & Rating & Quality & Informativeness \\
\midrule
$\pi_b$ (Chess.com) & 70.17 & 58.89 & 65.57 & 67.95 & 69.83 & 75.11 \\ 
$\pi_\theta$ (Ours) & 72.39 & 61.48 & 70.34 & 72.73 & 71.31 & 75.11 \\ \midrule
$\Delta$ & +2.22 & +2.59 & +4.77$^{**}$ & +4.78$^*$ & +1.48 & 0.00 \\
\bottomrule
\end{tabular}
\caption{Comparison of methods for qualitative expert-designed metrics. Metrics are converted to 100 from the original range to account for the difference in scale. Values are  averaged across all annotated puzzles. We perform Student's t-test and use $^*$ to mark $p < 0.05$, and $^{**}$ for $p < 0.01$.}
\label{table:method_comparison_1}
\end{table}


Our trained model recommends puzzles rated as slightly more enjoyable and slightly harder for the target player. The puzzles also receive somewhat higher calculation and pattern-recognition scores. Because these evaluations are generated by LLM judges calibrated on a small set of expert annotations, we treat the findings as preliminary.



\section{Conclusion and Future Work}
\label{sec:conclusion}

Our study leverages 1.5 billion puzzle-solving histories to learn the pedagogical value of chess puzzles and develop an automated puzzle selection system using offline reinforcement learning. 
Our offline policy evaluation and LLM-based qualitative analysis both suggest differences between the learned policy and Chess.com's existing policy. However, direct human evaluations are still needed to fully validate our system.
In the future, we hope to collaborate with our data providers to demonstrate the actual impact on chess learners by using our system to filter out low-quality chess puzzles and to investigate whether our approach has applications in other domains, such as math, language learning, or coding.


\section*{Acknowledgment}


During the period when this research was conducted, AN was supported in part by a Stanford HAI Hoffman-Yee grant and an NSF \#2112926 grant. 
NT was funded by a grant from FAR.AI.

Paul Terwilliger was our main collaborator from Chess.com, where he served as the director of AI at the time. Paul provided us with the dataset and participated in multiple research meetings to discuss potential uses of the data. We are extremely grateful for his contribution to the project. 

We also thank the rest of Chess.com for forming this collaboration with us and allowing us to explore different ways to empower chess learners all over the world.

Carissa Yip and Nicholas Tomlin served as our in-house chess experts. CY and NT co-designed the annotation schema used to rate each chess puzzle through an iterative process. CY helped recruit the expert chess annotators listed in Table \ref{tab:annotators}. We are grateful to the annotators for volunteering approximately 1.5 hours to complete a set of annotations on our chess puzzles. Each annotator was compensated with a Coupa Cafe gift card or an Amazon gift card.

\begin{table}[h]
\centering
\begin{tabular}{l l l}
\toprule
\textbf{Name} & \textbf{USCF Profile} & \textbf{Title} \\
\midrule
Alexander Su & \href{https://www.uschess.org/msa/MbrDtlMain.php?12857329}{12857329} & Expert \\
Robbie Selwyn & \href{https://www.uschess.org/msa/MbrDtlMain.php?17281090}{17281090} & N/A \\
Iris Zhou & \href{https://www.uschess.org/msa/MbrDtlMain.php?14467261}{14467261} & Expert \\
Tony Kukavica & \href{https://www.uschess.org/msa/MbrDtlMain.php?13853062}{13853062} & National Master \\
Justin Paul & \href{https://www.uschess.org/msa/MbrDtlMain.php?14323420}{14323420} & FIDE Master \\
Matt Larson & \href{https://www.uschess.org/msa/MbrDtlMain.php?14278511}{14278511} & International Master \\
Bryce Tiglon & \href{https://www.uschess.org/msa/MbrDtlMain.php?14230627}{14230627} & Grandmaster \\
Josiah Stearman & \href{https://www.uschess.org/msa/MbrDtlMain.php?14006506}{14006506} & Grandmaster \\
\bottomrule
\end{tabular}
\caption{Volunteer chess annotators who completed puzzle annotations, with their USCF profile (linked via their USCF member ID) and title.}
\label{tab:annotators}
\end{table}

\bibliography{main}
\bibliographystyle{rlj}

\appendix
\section{Appendix}
\subsection{Human Annotation Guidelines}
\label{app:annotation_guidelines}

This appendix describes the annotation protocol used to collect the expert ratings discussed in Section~\ref{sec:eval}. The goal of the annotation task is to measure whether puzzles recommended by our learned policy are of comparable or higher pedagogical quality than those served by Chess.com's existing tactics trainer. Each of the eight titled and expert-level annotators listed in Table~\ref{tab:annotators} was given a spreadsheet of puzzle URLs, together with the Chess.com Elo of the player each puzzle was originally recommended to and that player's puzzle rating. Annotators were asked to first attempt to solve each puzzle and record whether they solved it correctly, before rating it along the six criteria below and providing a short written description and justification (minimum length one sentence) for their ratings. Annotators were asked to budget approximately 1.5 hours to complete the full set of 30 puzzles, inclusive of reading these guidelines, and were compensated with a gift card.

\begin{figure}[h]
    \centering
    \begin{subfigure}[t]{0.47\textwidth}
        \centering
        \includegraphics[width=\textwidth]{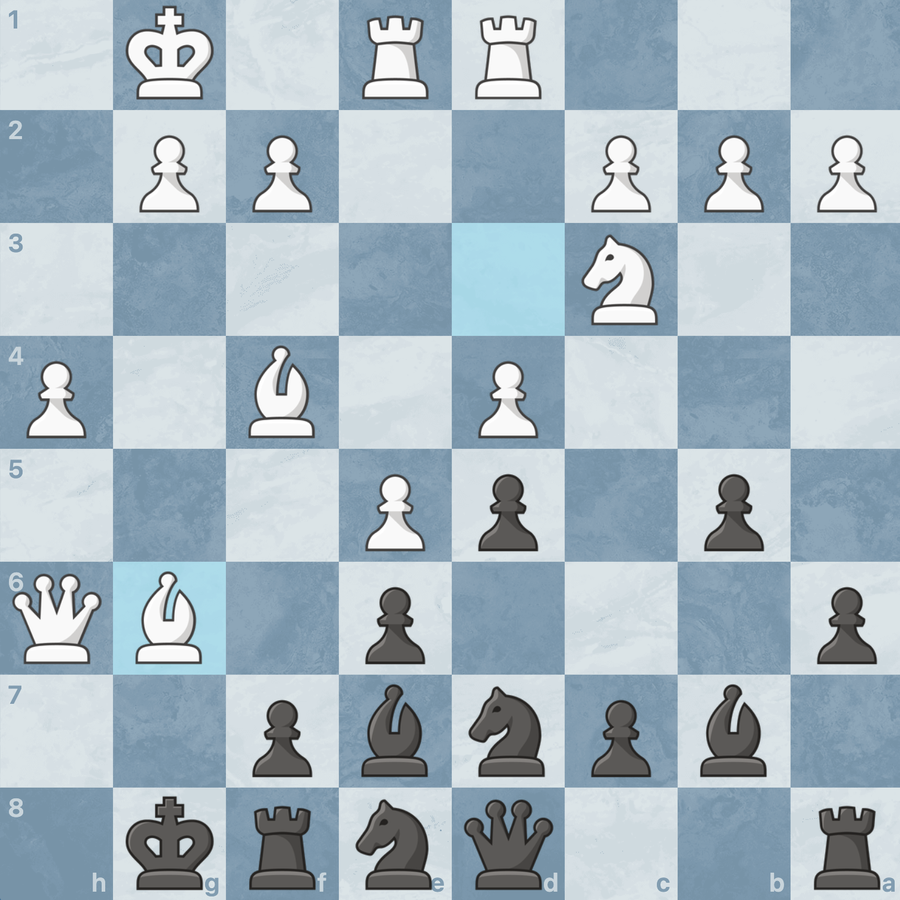}
        \caption{A \textit{calculation} score of 1: for instance, in the below position, Black is forced to take on g6 and play Ng7, which solves the puzzle.}
        \label{fig:appendix_calc_example}
    \end{subfigure}
    \hfill
    \begin{subfigure}[t]{0.47\textwidth}
        \centering
        \includegraphics[width=\textwidth]{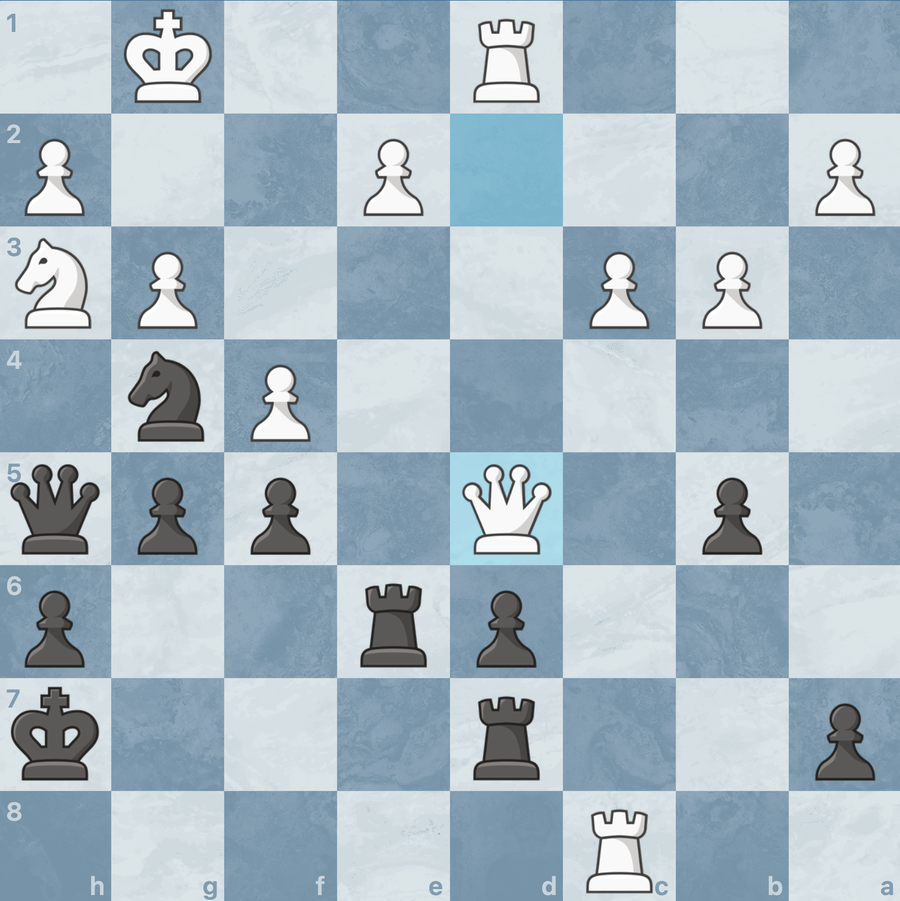}
        \caption{An \textit{informativeness} score of 1: for instance, in the below position, after we take on h3, the solution follows 2. Qg2 instead of taking on f5 (the most challenging line, which the player must have calculated before taking on h3).}
        \label{fig:appendix_info_example}
    \end{subfigure}
    \caption{Example puzzle positions used to illustrate the \textit{calculation} (Figure~\ref{fig:appendix_calc_example}) and \textit{informativeness} (Figure~\ref{fig:appendix_info_example}) criteria.}
    \label{fig:appendix_examples}
\end{figure}

\paragraph{Calculation} For each puzzle, annotators first judged whether it was primarily designed to test calculation, pattern recognition, both, or neither, and rated only the corresponding subcriterion below (marking the other as not applicable). \textit{Calculation} is scored on a 1--4 scale and reflects the extent to which a puzzle requires reasoning over long move sequences or lines with a high branching factor; puzzles that can be solved by identifying the single forcing move at each step, without look-ahead, receive a low score. Figure~\ref{fig:appendix_calc_example} shows an example rated a 1, in which every move in the solution is the only reasonable continuation available at that step.

\paragraph{Pattern recognition} \textit{Pattern recognition} is scored on a 1--3 scale and reflects how unusual or interesting a puzzle's tactical motif is, relative to common motifs (e.g., a standard smothered mate); a puzzle earns the highest score either for testing a genuinely unusual pattern or for giving a common pattern an unusual twist.

\paragraph{Informativeness} Scored on a 1--5 scale, \textit{informativeness} measures whether a puzzle's given solution follows the most challenging and instructive line rather than a secondary, less demanding continuation. Figure~\ref{fig:appendix_info_example} shows an example rated a 1, where the puzzle's solution avoids the most challenging continuation (which the solver must nonetheless calculate to justify an earlier move) in favor of an easier follow-up. Puzzles whose solutions instead follow the opponent's best defense, and whose logic remains clear once the solution is revealed, are rated highly.

\paragraph{Rating-appropriateness} Scored on a 1--5 scale, \textit{rating-appropriateness} captures the annotator's assessment of a puzzle's difficulty relative to the rating of the player it was recommended to. Annotators were asked to judge this criterion from the perspective of a player at the target rating, rather than their own.

\paragraph{Fun} Also scored on a 1--5 scale, \textit{fun} measures how enjoyable a puzzle would likely be for a player at the target rating.

\paragraph{Quality} \textit{Quality} is a holistic 1--5 score of the puzzle as a whole, independent of difficulty; the preceding criteria are intended to inform this judgment without strictly determining it. Figure~\ref{fig:rating_category} summarizes the rating range for each criterion.

\subsection{LLM Judge Prompt Fields}
\label{app:llm_prompt_fields}

Each LLM judge receives a system message and a DSPy signature. The system message identifies the expert annotator being imitated by their chess title, when applicable, and USCF Elo rating, and asks the model to apply the annotation guidelines together with that expert-level perspective. The signature contains the following three input fields:
\begin{itemize}
    \item \textit{Annotation guidelines}: the complete rubric for rating calculation, pattern recognition, informativeness, rating-appropriateness, quality, and fun.
    \item \textit{Target-user puzzle Elo}: the target user's current Chess.com puzzle Elo, which provides the reference skill level for rating-appropriateness and fun.
    \item \textit{Puzzle board}: an ASCII grid representation of the position, generated from the puzzle's FEN string, followed by a sentence describing the puzzle's first move and asking the solver to find the best continuation.
\end{itemize}
The signature's output field requests integer scores from 1 to 100 for all six categories in valid JSON format. DSPy also supplies selected expert-rated examples when executing the optimized judge.

\subsection{Annotation Questionnaire}
\label{app:annotation_questionnaire}

Annotators recorded their ratings in a shared spreadsheet, with one puzzle per row and one column per question. For each puzzle, the columns asked, verbatim:
\begin{itemize}
    \item \textit{What is the primary emphasis of this puzzle?} (Calculation / Pattern Recognition / Both / Neither)
    \item \textit{Calculation}: this puzzle requires the player to think through the entire solution before playing their first move. (1--4, or N/A)
    \item \textit{Pattern recognition}: this puzzle teaches useful or un-ordinary motifs, e.g. the motifs are not banal/overdone, or if they are, the puzzle puts a unique spin on them. (1--3, or N/A)
    \item \textit{Informativeness}: the solution goes down ``the right path'' to make sure the player understands the point of the puzzle. (1--5)
    \item \textit{Rating}: is this puzzle appropriate for someone rated at the target player's skill level? (1--5)
    \item \textit{Quality}: this is a high-quality puzzle, regardless of whether it is at the correct Elo level or not. (1--5)
    \item \textit{Fun}: I think this puzzle would be enjoyable and fun to solve by players at the target Elo. (1--5)
    \item \textit{Additional notes}: please write a 1--2 sentence justification of your ratings.
\end{itemize}
Table~\ref{tab:annotation_example} reproduces two rows from a completed annotation spreadsheet to illustrate how one annotator answered this questionnaire in practice.

\begin{table}[h]
\centering
\small
\begin{tabular}{l p{4.85cm} p{4.85cm}}
\toprule
\textbf{Field} & \textbf{Example 1} & \textbf{Example 2} \\
\midrule
Puzzle & \href{https://www.chess.com/puzzles/problem/1004456/practice}{\#1004456} & \href{https://www.chess.com/puzzles/problem/955974/practice}{\#955974} \\
Player Elo (Puzzle Elo) & 951 (1317) & 1361 (1615) \\ \midrule 
Solved correctly & Yes & Yes \\
Primary emphasis & Pattern Recognition & Both \\
Calculation & N/A & 1 -- easy to guess first move w/o seeing solution \\
Pattern recognition & 1 -- banal/boring pattern & 1 -- banal/boring pattern \\
Informativeness & 5 -- most informative & 1 -- least informative \\
Rating-appropriateness & 4 -- a bit too hard & 3 -- just right \\
Quality & 3 & 1 -- lowest quality \\
Fun & 4 & 2 \\ \midrule 
Annotator notes & \textit{``Rated as 4 difficulty because it is winning a piece not mate, and seeing the fork idea/long range check may be challenging for the rating.''} & \textit{``The puzzle if done properly is about right in terms of difficulty. As is, it is bad. The puzzle does not even give a line, which is horrifically bad, and makes it easy to guess the correct answer without full understanding/calculation.''} \\
\bottomrule
\end{tabular}
\caption{Two example rows from a completed annotation spreadsheet, showing how one annotator answered the questionnaire in Appendix~\ref{app:annotation_questionnaire} for two puzzles.}
\label{tab:annotation_example}
\end{table}

\subsection{Selecting Illustrative Examples}
\label{app:example_selection}

To illustrate how ratings can differ across puzzles, and to qualitatively showcase the difference between puzzles recommended by our system and the original system, we selected a second pair of example puzzles, shown in Table~\ref{tab:pattern_example}. As with Table~\ref{tab:annotation_example}, we restricted our search to the subset of puzzles that all eight annotators independently rated (the ``preference learning'' puzzles described in Section~\ref{sec:eval}). Within this subset, we computed each puzzle's average pattern recognition score (excluding annotators for whom the criterion did not apply) and selected one puzzle near the top and one near the bottom of this ranking, from our policy and from Chess.com's existing system respectively.

The first example (a puzzle recommended by our policy) received the maximum pattern recognition score from every annotator who rated it: all described the position's combination of moves as an unusual or non-obvious way to reach the tactical idea, distinct from a puzzle that merely tests whether the solver knows a standard pattern.

The second example (a puzzle served by Chess.com's existing system) had a lower average pattern recognition score, with annotators ranging from ``banal/boring pattern'' to ``interesting and useful'' for the same puzzle. Notably, its holistic quality score remained comparable to the first example's, since most annotators still considered it a reasonably solid puzzle overall; only the pattern recognition criterion, and the accompanying free-text notes, surfaced the disagreement over whether its underlying motif was novel or overdone. This is precisely the gap we selected these examples to illustrate: a puzzle can score well on quality while still being seen by some experts as testing an unremarkable, overdone pattern.

\begin{table}[h]
\centering
\small
\begin{tabular}{l p{4.5cm} p{4.5cm}}
\toprule
\textbf{Field} & \textbf{Example 1 (Ours)} & \textbf{Example 2 (Chess.com)} \\
\midrule
Puzzle & \href{https://www.chess.com/puzzles/problem/760014/practice}{\#760014} & \href{https://www.chess.com/puzzles/problem/1072600/practice}{\#1072600} \\
Player Elo (Puzzle Elo) & 2175 (3053) & 951 (1317) \\ \midrule 
Solved correctly & Yes & Yes \\
Primary emphasis & Both & Calculation \\
Calculation & 4 -- requires full understanding of solution to solve & 2 \\
Pattern recognition & 3 -- interesting and useful & 1 -- banal/boring pattern \\
Informativeness & 5 -- most informative & 4 \\
Rating-appropriateness & 3 -- just right & 2 -- a bit too easy \\
Quality & 5 -- highest quality & 3 \\
Fun & 5 -- most fun & 2 \\ \midrule 
Annotator notes & \textit{``A+ for fun and useful. Also, b5+ and Rd5+ is a particularly unusual combination which requires calculation... as well as nice checking pattern recognition. Good puzzle!''} & \textit{``A bit of a boring problem which forces White to find all checks. It is possible that White forgets to give the checks and creates the mating threat with 1.Qf7, which at this level where checks are the first thing to examine the problem isn't too difficult.''} \\
\bottomrule
\end{tabular}
\caption{Two example rows from a completed annotation spreadsheet, selected to contrast a puzzle with a unanimously high pattern recognition score (Example 1) against a puzzle with a lower, more disputed pattern recognition score (Example 2), as discussed in Appendix~\ref{app:example_selection}.}
\label{tab:pattern_example}
\end{table}

Table~\ref{tab:example_distribution} shows the full pattern recognition and quality distributions across all eight annotators for both puzzles, rather than the mean alone, so that the degree of agreement (or disagreement) on each criterion is directly visible.

\begin{table}[h]
\centering
\begin{tabular}{c || c c | c c}
\toprule
\textbf{Annotator} & \multicolumn{2}{c|}{\textbf{Pattern Recognition}} & \multicolumn{2}{c}{\textbf{Quality}} \\
 & Ours & Chess.com & Ours & Chess.com \\
\midrule
A & 3 & 2 & 3 & 4 \\
B & N/A & 2 & 5 & 4 \\
C & N/A & N/A & 4 & 5 \\
D & 3 & 1 & 5 & 3 \\
E & 3 & 3 & 3 & 4 \\
F & 3 & 3 & 5 & 5 \\
G & N/A & N/A & 5 & 4 \\
H & 3 & 3 & 4 & 4 \\
\midrule
Mean & 3.00 & 2.33 & 4.25 & 4.12 \\
\bottomrule
\end{tabular}
\caption{Pattern recognition and quality scores given by each of the eight annotators to the two puzzles in Table~\ref{tab:pattern_example} (``Ours'' is \#760014 and ``Chess.com'' is \#1072600; N/A denotes an annotator for whom the pattern recognition criterion did not apply), with rows aligned so that the same letter denotes the same annotator across all four columns. While the two puzzles' quality scores are similar on average, their pattern recognition scores diverge, with \#1072600 receiving both the single lowest score (a ``banal/boring pattern'' rating) and a wider spread of opinions overall.}
\label{tab:example_distribution}
\end{table}

\end{document}